\documentclass{article}
\PassOptionsToPackage{pagebackref=true}{hyperref}
\usepackage[preprint,nonatbib]{corl_2026}
\usepackage[square,numbers,sort&compress]{natbib}
\usepackage{enumitem}

\usepackage{times}
 \usepackage[printonlyused]{acronym}
\usepackage[export]{adjustbox} %
\usepackage{algorithm}
\usepackage{algorithmicx}
\usepackage{algpseudocode}
\algrenewcommand\algorithmicindent{0.8em}%
\usepackage{amssymb}
\usepackage{amsmath}
\usepackage{array}
\usepackage{bbm}
\usepackage{booktabs}
\usepackage{caption}
\usepackage{float}
\usepackage{appendix} 
\usepackage{graphicx}
\usepackage[pagebackref=true]{hyperref}
\usepackage{cleveref}
\usepackage{lipsum}
\usepackage{microtype}
\usepackage{multicol}
\usepackage{multirow}
\usepackage{nicefrac}
\usepackage{relsize}
\usepackage{siunitx}
\usepackage{subcaption}
\usepackage{todonotes}
\usepackage{wrapfig}
\usepackage{tikz}
\usepackage{verbatim}
\usepackage{xspace}
\usepackage{enumitem}

\usepackage{amsthm}
\algnewcommand{\IfThenElse}[3]{%
  \State \algorithmicif\ #1\ \algorithmicthen\ #2\ \algorithmicelse\ #3}
 \algnewcommand{\IfThen}[2]{%
  \State \algorithmicif\ #1\ \algorithmicthen\ #2}

\usepackage[utf8]{inputenc} %
\usepackage[T1]{fontenc}    %
\usepackage{url}            %
\usepackage{booktabs}       %
\usepackage{amsfonts}       %
\usepackage{nicefrac}       %
\usepackage{microtype}      %

\definecolor{mydarkblue}{rgb}{0,0.08,0.45}
\hypersetup{ %
    pdftitle={},
    pdfauthor={},
    pdfsubject={},
    pdfkeywords={},
    pdfborder=0 0 0,
    pdfpagemode=UseNone,
    colorlinks=true,
    linkcolor=mydarkblue,
    citecolor=mydarkblue,
    filecolor=mydarkblue,
    urlcolor=mydarkblue,
    pdfview=FitH}

\algnewcommand{\algorithmicforeach}{\textbf{for each}}
\algdef{SE}[FOR]{ForEach}{EndForEach}[1]
  {\algorithmicforeach\ #1\ \algorithmicdo}%
  {\algorithmicend\ \algorithmicforeach}%
\newcommand{\pushright}[1]{\ifmeasuring@#1\else\omit\hfill$\displaystyle#1$\fi\ignorespaces}

\usetikzlibrary{tikzmark}
\usetikzlibrary{calc}
\newcommand{\ALGtikzmarkcolor}{black}%
\newcommand{\ALGtikzmarkextraindent}{2pt}%
\newcommand{\ALGtikzmarkverticaloffsetstart}{-.5ex}%
\newcommand{\ALGtikzmarkverticaloffsetend}{-.5ex}%
\makeatletter
\newcounter{ALG@tikzmark@tempcnta}

\newcommand\ALG@tikzmark@start{%
    \global\let\ALG@tikzmark@last\ALG@tikzmark@starttext%
    \expandafter\edef\csname ALG@tikzmark@\theALG@nested\endcsname{\theALG@tikzmark@tempcnta}%
    \tikzmark{ALG@tikzmark@start@\csname ALG@tikzmark@\theALG@nested\endcsname}%
    \addtocounter{ALG@tikzmark@tempcnta}{1}%
}

\def\ALG@tikzmark@starttext{start}
\newcommand\ALG@tikzmark@end{%
    \ifx\ALG@tikzmark@last\ALG@tikzmark@starttext
    \else
        \tikzmark{ALG@tikzmark@end@\csname ALG@tikzmark@\theALG@nested\endcsname}%
        \tikz[overlay,remember picture] \draw[\ALGtikzmarkcolor] let \p{S}=($(pic cs:ALG@tikzmark@start@\csname ALG@tikzmark@\theALG@nested\endcsname)+(\ALGtikzmarkextraindent,\ALGtikzmarkverticaloffsetstart)$), \p{E}=($(pic cs:ALG@tikzmark@end@\csname ALG@tikzmark@\theALG@nested\endcsname)+(\ALGtikzmarkextraindent,\ALGtikzmarkverticaloffsetend)$) in (\x{S},\y{S})--(\x{S},\y{E});%
    \fi
    \gdef\ALG@tikzmark@last{end}%
}

\apptocmd{\ALG@beginblock}{\ALG@tikzmark@start}{}{\errmessage{failed to patch}}
\pretocmd{\ALG@endblock}{\ALG@tikzmark@end}{}{\errmessage{failed to patch}}

\newcommand{\x}{\mathbf{x}}

\newcommand{\y}{\mathbf{y}}

\newcommand{\gauss}{\mathcal{N}}

\newcommand{\reward}{r}

\newcommand{\states}{\mathcal{S}}
\newcommand{\actions}{\mathcal{A}}

\newcommand{\eye}{\mathbf{I}}

\newcommand{\bo}{\mathbf{o}}
\newcommand{\bs}{\mathbf{s}}
\newcommand{\ba}{\mathbf{a}}

\newtheorem*{theorem-nonlabeled}{Theorem}

\newcommand{\expectation}{\mathbb{E}}

\newcommand{\given}{\:\vert\:}

\newcommand{\discountFactor}{\gamma}

\newcommand{\recipeExpert}{\textsc{Expert-only}\xspace}   %
\newcommand{\recipeFull}{\textsc{Full-FT}\xspace}         %
\newcommand{\recipeFrozenLM}{\textsc{Frozen-LM}\xspace}   %
\newcommand{\recipeLoRA}{\textsc{LoRA}\xspace}            %
\newcommand{\recipeLoRALM}{\textsc{LoRA-LM}\xspace}       %

\acrodef{VLA}{Vision-Language-Action model}

\acrodef{AGI}{artificial general intelligence}
\acrodef{ANOVA}[ANOVA]{Analysis of Variance\acroextra{, a set of
  statistical techniques to identify sources of variability between groups}}
\acrodef{ANN}{artificial neural network}
\acrodef{API}{application programming interface}
\acrodef{CACLA}{continuous actor critic learning automaton}
\acrodef{cGAN}{conditional generative adversarial network}
\acrodef{CMA}{covariance matrix adaptation}
\acrodef{COM}{centre of mass}
\acrodef{CTAN}{\acroextra{The }Common \TeX\ Archive Network}
\acrodef{DDPG}{deep deterministic policy gradient}
\acrodef{DeepLoco}{deep locomotion}
\acrodef{DOI}{Document Object Identifier\acroextra{ (see
    \url{http://doi.org})}}
\acrodef{DPG}{deterministic policy gradient}
\acrodef{DQN}{deep Q-network}
\acrodef{DRL}{deep reinforcement learning}
\acrodef{DYNA}{DYNA}
\acrodef{EOM}{Equations of motion}
\acrodef{EPG}{expected policy gradient}
\acrodef{FDR}{future discounted reward}
\acrodef{FSM}{finite state machine}
\acrodef{GAE}{generalized advantage estimation}
\acrodef{GAIfO}{generative adversarial imitation from observation}
\acrodef{GAN}{generative adversarial network}
\acrodef{GPS}[GPS]{Graduate and Postdoctoral Studies}
\acrodef{HLC}{high-level controller}
\acrodef{HLP}{high-level policy}
\acrodef{HRL}{hierarchical reinforcement learning}
\acrodef{KLD}{Kullback-Leibler divergence}
\acrodef{LLC}{low-level controller}
\acrodef{LLP}{low-level policy}
\acrodef{MARL}{Multi-Agent Reinforcement Learning}
\acrodef{MBAE}{model-based action exploration}
\acrodef{MPC}{model predictive control}
\acrodef{MDP}{Markov Decision Processes}
\acrodef{DSRL}{Diffusion Steering Reinforcement Learning}
\acrodef{SFT}{Supervised fine-tuning}
\acrodef{MSE}{mean squared error}
\acrodef{MultiTasker}{controller that learns multiple tasks at the same time}
\acrodef{Parallel}{randomly initialize controllers and train in parallel}
\acrodef{PD}{proportional derivative}
\acrodef{PDF}{Portable Document Format}
\acrodef{PLAiD}{Progressive Learning and Integration via Distillation}
\acrodef{PPO}{proximal policy optimization}
\acrodef{PTD}{positive temporal difference}
\acrodef{RBF}{radial basis function}
\acrodef{ReLU}{rectified linear unit}
\acrodef{RCS}[RCS]{Revision control system\acroextra{, a software
    tool for tracking changes to a set of files}}
\acrodef{RL}{reinforcement learning}
\acrodef{SGD}{stochastic gradient descent}
\acrodef{Scratch}{randomly initialized controller}
\acrodef{SIMBICON}{SIMple BIped CONtroller}
\acrodef{SMBAE}{stochastic model-based action exploration}
\acrodef{SVG}{stochastic value gradients}
\acrodef{SVM}{support vector machine}
\acrodef{TCN}{time contrastive learning}
\acrodef{TD}{temporal difference}
\acrodef{TL}{transfer learning}
\acrodef{terrainRL}{terrain adaptive locomotion}
\acrodef{TLX}[TLX]{Task Load Index\acroextra{, an instrument for gauging
  the subjective mental workload experienced by a human in performing
  a task}}
\acrodef{TRPO}{trust region policy optimization}
\acrodef{UBC}{University of British Columbia}
\acrodef{UCB}{upper confidence bound}
\acrodef{UI}{user interface}
\acrodef{UML}{Unified Modelling Language\acroextra{, a visual language
    for modelling the structure of software artefacts}}
\acrodef{URDF}{unified robot description format}
\acrodef{URL}{Unique Resource Locator\acroextra{, used to describe a
    means for obtaining some resource on the world wide web}}
\acrodef{W3C}[W3C]{\acroextra{the }World Wide Web Consortium\acroextra{,
    the standards body for web technologies}}    
\acrodef{XML}{Extensible Markup Language}

\title{Finetuning VLAs for Dexterous Manipulation: The Right Checkpoint Is All You Need}

\author{
  \bfseries Artur Kuramshin\textsuperscript{1,2}, \; Charlotte Cloutier\textsuperscript{1,2}, \; Michael Przystupa\textsuperscript{1,2} \\
  \bfseries Kirsty Ellis\textsuperscript{1,2}, \; Mohammad S. Akhter\textsuperscript{3}, \; Glen Berseth\textsuperscript{1,2} \\
  \textsuperscript{1}Mila --- Quebec AI Institute, Montr\'eal, Canada \quad
  \textsuperscript{2}Universit\'e de Montr\'eal, Canada \\
  \textsuperscript{3}Renesas Electronics Canada Ltd.
}

\begin{document}
\maketitle

\begin{abstract}

Vision-Language-Action models have demonstrated strong manipulation capabilities. A limitation of VLA models is that most manipulation datasets focus on parallel-jaw gripper systems, making transfer to multi-finger hands challenging, particularly as teleoperation datasets are harder to collect with multi-fingered robots. 
In this work, we study how to best adapt VLAs to dexterous hands by pairing supervised fine-tuning (SFT) with diffusion-steering reinforcement learning (DSRL).
Using $\pi_{0.5}$ and a LEAP hand, we evaluate five fine-tuning recipes across two simulation tasks using only a handful of demonstrations.
Our empirical analysis yields three key insights: (1) SFT success is poorly correlated with post-DSRL success, meaning the best SFT checkpoint is rarely the most effective to steer; (2) freezing the vision-language model (VLM) and tuning only the action expert produces the most steerable policy for both training and out-of-distribution placements; and (3) low-rank adaptation (LoRA) is the least steerable approach. Consequently, we recommend an adaptation recipe that freezes the VLM, fully trains the action expert, and eschews SFT rollout success as a checkpoint selection metric. 
\end{abstract}


\section{Introduction}
\label{sec:intro}

Vision-Language-Action models (VLA) \cite{black2024pi0,intelligence2025pi05,bjorck2025gr00t,kim2024openvla} are largely pretrained on cross-embodiment corpora dominated by parallel-jaw grippers \cite{o2024openXembodiment,Khazatsky2024Droid}. This confines them to tasks that a two-finger gripper can solve, which excludes many tasks involving tool use and reorientation of the hand \cite{zhang2025dexterousmanufacturing,an2025dexterousILsurvey}. Because multi-fingered hands are scarce in existing pretraining data \cite{zhang2026dexora,luo2026being}, deploying a VLA on a novel dexterous embodiment necessitates fine-tuning. Furthermore, this adaptation must succeed on a limited number of demonstrations due to the high data-collection cost associated with dexterous teleoperation \cite{TaoT-RSS-25,arcap2025,guzey2025dexteritysmartlensesmultifingered}. In this work, we investigate how to adapt VLA models for tasks with multi-fingered hands on limited data.

How to fine-tune a VLA, meaning which of its parameters to update and at what rate, is already contested: some report that freezing the language model or the vision encoder degrades adaptation \cite{ferchau2026efficiency,kim2024openvla}, others keep action-pathway gradients out of the backbone by default \cite{nvidia2025gr00tn15,driess2025knowledge}. These conclusions were reached by evaluating adaptation solely based on the supervised fine-tuning (SFT) checkpoint with the highest rollout success. When adaptation pipelines culminate in latent-noise steering \cite{wagenmaker2025DSRL,yu2026support}, the SFT checkpoint merely serves as an initialization. The critical factor is instead whether the frozen model's noise space admits an actor that succeeds and generalizes beyond its training states.

In this paper, we investigate five SFT recipes to determine which, if any, reliably supports dexterous manipulation transfer when followed by latent steering. Our recipes vary which parts of the VLA we train (action expert only, plus the vision pathway, or the full model) and whether we update the language model and action expert fully or via LoRA \cite{hu2022lora} adapters. Our analysis applies these combinations to $\pi_{0.5}$ \cite{intelligence2025pi05} on a 16-DoF LEAP hand \cite{shaw2023leap} across two simulation tasks. We first compare VLA fine-tuning recipes as initialization for latent steering. Based on our empirical analysis, we provide a concrete recommendation: when SFT is followed by latent steering, freeze the VLM, train the action expert in full, and avoid using SFT rollout success for recipe or checkpoint selection. Furthermore, the choice of checkpoint mattered less than the choice of recipe. Since steerability declines over the course of SFT while motion quality improves, we suggest selecting the earliest checkpoint that qualitatively reproduces the demonstrated skill as the initialization for steering.

\section{Related Work}
\label{sec:related}

\textbf{Vision-language-action models.} VLAs adapt a pretrained vision-language model on robot data so that it emits low-level actions
\cite{black2024pi0,kim2024openvla,octo_2023}. The prevailing design consists of: a VLM backbone encodes images and language, and a smaller
generative action expert decodes a chunk of continuous actions from the backbone's features by diffusion or flow matching
\cite{black2024pi0,intelligence2025pi05,bjorck2025gr00t,fang2026molmoact2}. The VLA approach has been extended to dexterous manipulation by training from scratch on new data sources---egocentric human video and large multi-fingered corpora
\cite{nvidia2026gr00tn17,luo2026being,li2025vitra}. We take a complementary route: adapting an existing gripper-pretrained VLA to a multi-fingered hand from a few dozen demonstrations.

\textbf{Fine-tuning VLAs.} How to fine-tune a VLA for a new setup is an open question, particularly as prior research shows mixed results. OpenVLA \cite{kim2024openvla} and Ferchau et al.\ \cite{ferchau2026efficiency} find rank-32 LoRA on the backbone and action expert competitive with full fine-tuning, but Ferchau et al.\ also find that freezing the VLM, or restricting the vision encoder to LoRA, degrades performance sharply. GR00T N1.5 \cite{nvidia2025gr00tn15} and knowledge
insulation \cite{driess2025knowledge} reach the opposite default, keeping action-pathway gradients out of the language model. OpenVLA-OFT \cite{kim2025fine} studies the decoding scheme, action representation, and objective rather than deciding which parameters to update, and, like the others, selects the checkpoints by the success of the SFT rollout.

\textbf{Improving VLAs with reinforcement learning.} Online improvement of a fine-tuned VLA takes two forms. One updates the model's weights, with stabilization against forgetting \cite{intelligence2025pi,guo2025conrft,lu2025irevla}. Others leave the model frozen and learn a small policy over its generative latent: Diffusion Steering Reinforcement Learning (DSRL) \cite{wagenmaker2025DSRL} steers the initial noise of a diffusion or flow policy. Residual flow steering \cite{RFS2026Su} adds a learned action residual for dexterous tasks. We adopt DSRL as the endpoint because it needs no distillation and never touches the fine-tuned weights, so differences between recipes act only through the frozen model. These works take the base policy as given; none asks how to fine-tune it, or which checkpoint to steer from. The closest work to ours is SCORE \cite{yu2026support}, which frames steering as support-constrained RL and uses it to improve real-world multi-fingered policies through simulation. SCORE's data ablation finds that retry and play data leave base-policy success unchanged but substantially raise post-steering success, suggesting the prior should be prepared for steering rather than zero-shot performance. SCORE establishes this along the data axis for a point-cloud flow policy trained from scratch; we establish it along the fine-tuning-recipe axis for an image-based VLA with large-scale pretraining.

\section{Studying VLA Adaptation Recipes for Steering}
\label{sec:method} 

Our goal is to determine which SFT recipe produces the best policy when adaptation of a dual-system VLA is followed by latent-noise steering. We first provide preliminary information to put our work into context, define the design decisions a recipe makes and the five recipes that isolate them (\cref{sec:method:decisions}), and then describe how each recipe is executed on $\pi_{0.5}$ (App.~\ref{app:recipes}) and how the resulting checkpoints are steered with DSRL and evaluated (\cref{sec:method:dsrl}).
\subsection{Preliminary}\label{sec:background}

\textbf{The} $\mathbf{\pi_{0.5}}$ \textbf{model.} We study $\pi_{0.5}$ \cite{intelligence2025pi05}, a dual-system VLA in which a PaliGemma backbone \cite{beyer2024paligemma} encodes images, language and the binned proprioceptive state, and a flow-matching action expert produces chunks of $H=32$ actions \cite{zhao2023act}. A new embodiment affects the VLA's state through the language model, and its actions through the expert's linear projections (see App.~\ref{app:model}).

\textbf{Diffusion steering RL.} The expert maps Gaussian noise $\mathbf{A}^{0}\sim\gauss(0,\eye)$ to an action chunk by integrating a learned velocity field conditioned on the observation $\bo_t$ \cite{lipman2023flow,black2024pi0}. SFT continues this training with gradients restricted to a subset $\theta_{\text{train}}\subseteq\theta$ of the weights. The chosen parameter subset and learning rate is how we define an SFT recipe. At inference, the chunk is a deterministic function $\mathbf{A}_t=f_\theta(\bo_t,\mathbf{A}^0)$. DSRL \cite{wagenmaker2025DSRL} exploits this: treating the task as a \ac{MDP} with reward $\reward_t$ and discount $\discountFactor$, it leaves $\theta$ frozen and learns a policy $\pi_\phi(\mathbf{w}\given\bo_t)$ over the noise $\mathbf{w}=\mathbf{A}^0$, crediting each executed chunk's reward to the $\mathbf{w}$ that produced it.

\subsection{Adaptation design decisions}
\label{sec:method:decisions}

We investigate the adaptation of a pretrained VLA to an embodiment absent from its pretraining data, utilizing a few dozen teleoperated demonstrations per task. Our pipeline consists of SFT on the demonstration data, followed by online improvement via DSRL. We isolate four key design decisions: updating the language model (H1), updating the vision encoder (H2), and updating the backbone/action expert either fully or through low-rank adapters (H3, H4). We construct five distinct recipes to evaluate these axes:

\begin{itemize}
    \item \textbf{\recipeExpert}: The action expert and its interface to the backbone are trained; the entire VLM remains frozen.
    \item \textbf{\recipeFrozenLM}: Extends \recipeExpert\ by unfreezing the vision encoder and multimodal projector, isolating the impact of the vision pathway (H2).
    \item \textbf{\recipeFull}: All parameters are trained end-to-end. Contrasting this with \recipeExpert\ and \recipeFrozenLM\ isolates the effect of language model adaptation (H1).
    \item \textbf{\recipeLoRA}: Applies rank-32 LoRA \cite{hu2022lora} to the language model and action expert, while fully training the vision encoder, multimodal projector, and action projections.
    \item \textbf{\recipeLoRALM}: Matches \recipeFrozenLM, but adds rank-32 LoRA to the language model. This isolates language model adaptation with fixed representations (H1) and contrasts with \recipeLoRA\ regarding full vs. adapter-based expert training (H4). Comparing this against \recipeFull\ isolates the update parameterization (H3).
\end{itemize}

The recipes use a common peak learning rate (ten times higher for LoRA parameters \cite{schulman2025lora}) and all other training settings are shared. App.~\ref{app:recipes} lists the exact parameter groups and hyperparameters.

\subsection{Steering with DSRL}
\label{sec:method:dsrl}

We use the large-chunk variant of \ac{DSRL} \cite{wagenmaker2025DSRL}: the actor outputs one noise vector $\mathbf{w}\in[-1,1]^{32}$, repeated across the $H=32$ chunk positions, and the frozen checkpoint decodes it. The actor observes a low-resolution image and the proprioceptive state through its own encoder. Every RL hyperparameter is shared across recipes, checkpoints, and tasks (App.~\ref{app:dsrl}).

\textbf{Checkpoint selection.} Supervised fine-tuning produces a sequence of checkpoints, and the prevailing practice is to deploy the one with the highest rollout success \cite{kim2025fine}. Whether that heuristic also picks the best checkpoint to steer is unknown, so we choose four checkpoints per recipe and task, from early to late training, and record each one's SFT-time success alongside its post-DSRL success. SFT-time success is measured under both i.i.d. and repeated noise App.~\ref{app:eval}.

\begin{figure}[ht]
\centering
\includegraphics[width=0.7\linewidth]{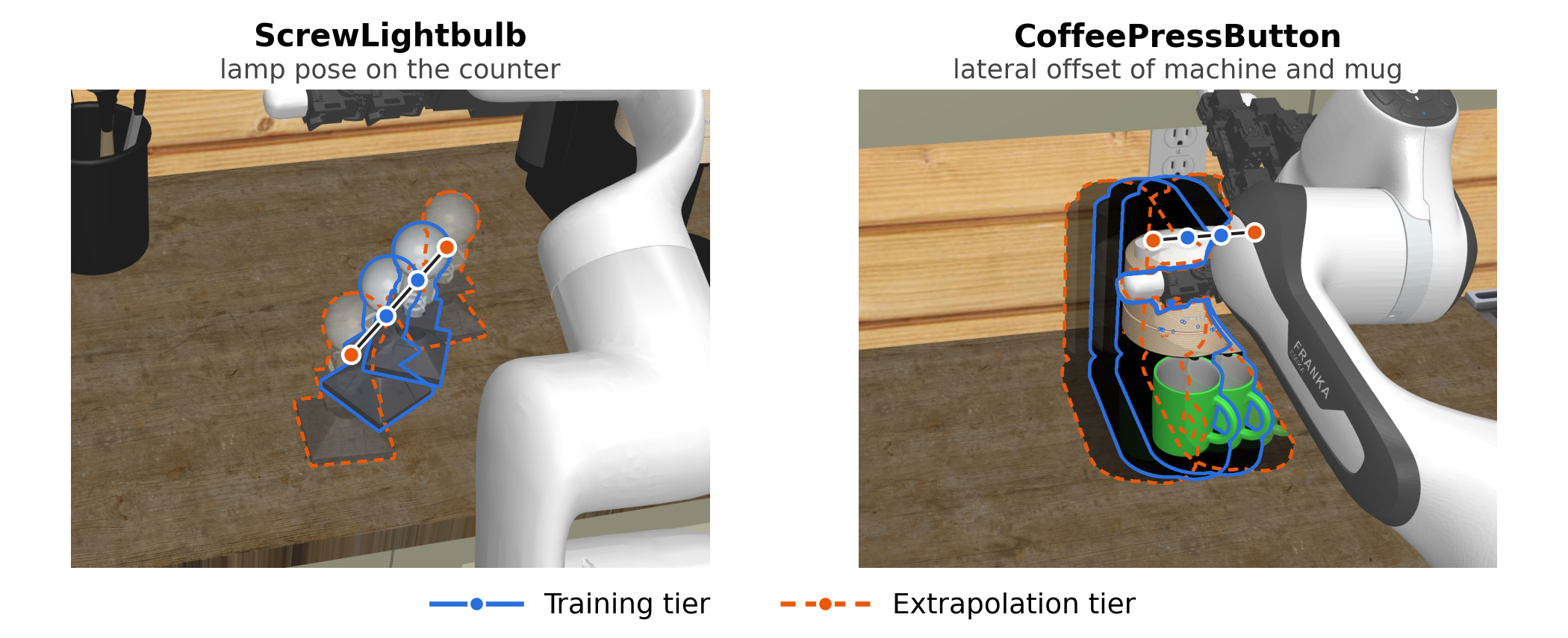}
\caption{\textbf{Tasks.} Each task evaluates spatial generalization by varying a single object placement along a defined axis. Left: \textsc{ScrewLightbulb}, varying the lamp pose on the counter. Right: \textsc{CoffeePressButton}, varying the lateral offset of the coffee machine. Policies are evaluated on both training placements (used during DSRL) and extrapolation placements (unseen spatial offsets).}
\label{fig:placements}
\end{figure}

\textbf{Evaluation setup.} Each task varies the placement of one object along a single axis
(\cref{fig:placements}). DSRL trains on two placements, alternating between them on reset (\textbf{training} placements); every policy is also scored on two placements outside the span the actor sees during training (\textbf{extrapolation} placements). Both evaluation conditions are scored before and after RL.

\section{Experimental Results}
\label{sec:results}

\begin{figure}[ht]
\centering
\includegraphics[width=\textwidth]{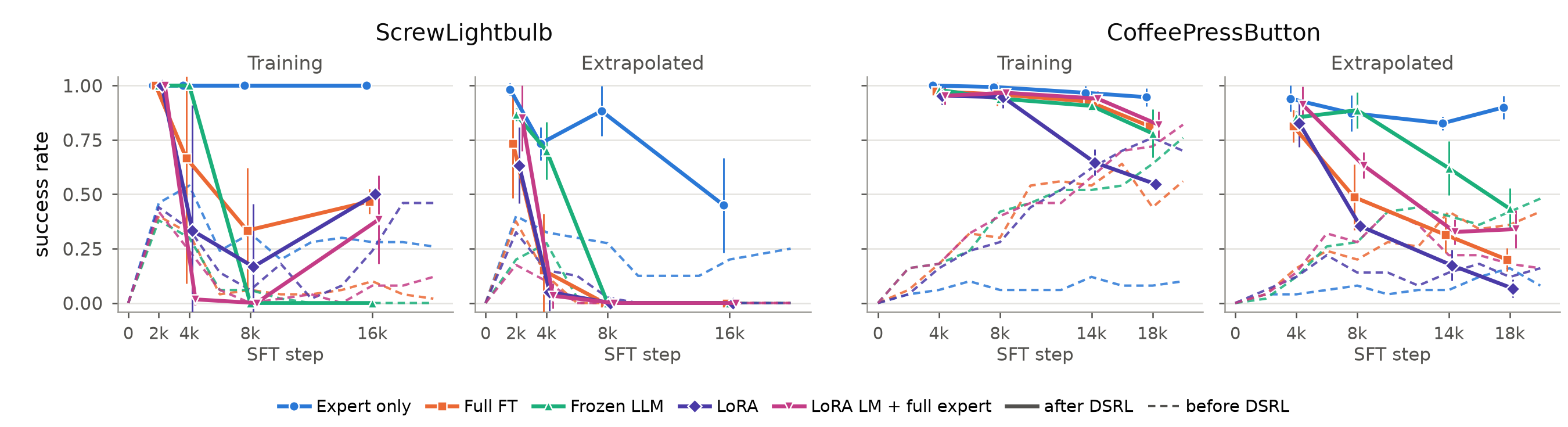}
\caption{\textbf{Success before (dashed) and after (solid) steering, by SFT step.} Both pre- and post-DSRL evaluations utilize repeated noise sampling to measure the marginal success rate over the latent space searched by the actor.}
\label{fig:main}
\end{figure}

\textbf{Setup.} The new embodiment is a 16-DoF LEAP hand \cite{shaw2023leap}, absent from $\pi_{0.5}$'s pretraining, on two custom tasks in the RoboCasa \cite{robocasa2024} simulation framework. \textsc{CoffeePressButton} requires pressing a coffee machine's button with a finger and \textsc{ScrewLightbulb} requires grasping a bulb and twisting it into a lamp socket. Demonstrations were teleoperated in simulation (App.~\ref{app:tasks}). We steer SFT steps 2k/4k/8k/16k on \textsc{ScrewLightbulb} and 4k/8k/14k/18k on \textsc{CoffeePressButton}. We report success on the training and extrapolation placements of \cref{sec:method:dsrl}, before RL (repeated noise) and after RL (mean of three seeds). \Cref{fig:main} shows each configuration evaluated and App.~\ref{app:full-results} shows the exhaustive table.

\vspace{-6pt}

\textbf{SFT-time success is a poor predictor of post-steering success.} Across all 20 recipe--checkpoint configurations,  pre-RL success (for both i.i.d. and repeated noise) exhibits no significant positive correlation with post-DSRL success on \textsc{ScrewLightbulb} 
and demonstrates anti-correlation on \textsc{CoffeePressButton}.
Consequently, selecting checkpoints based on standard SFT rollout success yields suboptimal steering initializations. While SFT success measures the policy's capacity to generate solutions via its own sampling, DSRL requires the noise space to contain a robust, generalizable success region.

\textbf{Adapting only the action expert is sufficient and yields the highest steerability.} \recipeExpert\ is the only recipe that steers to consistent success across every evaluated configuration, achieving 95--100\% on training placements for both tasks. Conversely, the three recipes that update the language model consistently perform the worst.
Neither LoRA variant protects against the degradation from updating the language model. Unfreezing the vision encoder (\recipeFrozenLM) matches \recipeExpert\ at early checkpoints and gives the highest SFT-time success of any recipe on \textsc{CoffeePressButton}, but it collapses at later checkpoints. \recipeExpert is also the only recipe that remains steerable at later checkpoints: at 8k on \textsc{ScrewLightbulb} it reaches 100\% on the training placements and 88\% on extrapolation, while every other recipe is at or below 33\%.

\section{Limitations}
\label{sec:limitations}

Our comparison has three limitations. First, each recipe was fine-tuned once, so the three RL seeds capture the variance of steering but not of SFT; confirming the effects requires repeated SFT runs. Second, the study is in simulation, and steering exploits it: the earliest \textsc{ScrewLightbulb} checkpoints turn the bulb by using the palm rather than grasping it (\Cref{fig:filmstrip}), so early-checkpoint success overstates usable skill and real-robot validation is needed. Third, SFT-time success does not predict the steering outcome, and an offline metric is the natural next step.

\section*{Acknowledgments}

We acknowledge the use of Claude Sonnet 5.5 (Anthropic) for assistance with editing and refining the written presentation of this work. All scientific content, methodology, and conclusions remain the original contributions of the authors.

We want to acknowledge funding support from Natural Sciences and Engineering Research Council of Canada, Fonds de recherche du Québec and The Canadian Institute for Advanced Research (CIFAR) and compute support from Digital Research Alliance of Canada, Mila IDT, and NVidia. This work was supported in part by Renesas Electronics through a research partnership on dexterous manipulation.

\bibliographystyle{plainnat}
\bibliography{references}

\newpage
\begin{appendices}

\section{The $\pi_{0.5}$ model and flow-matching action generation}
\label{app:model}
 
\textbf{Architecture.} The backbone is a 3B-parameter PaliGemma VLM \cite{beyer2024paligemma}: a SigLIP vision encoder, a multimodal projector, and a Gemma language model. The action expert is a smaller transformer that attends to the backbone's key--value cache and is conditioned on the flow timestep through adaptive RMSNorm \cite{NEURIPS2019_rmsnorm} modulation layers in every block. Observations consist of camera images, a language instruction and the proprioceptive state. Actions are 32-dimensional vectors (shorter action spaces are zero-padded) predicted in chunks of $H=32$ steps \cite{zhao2023act}; the noisy action chunk enters and leaves the expert through linear action projections.
 
\textbf{Flow matching.} Starting from $\mathbf{A}^{0}\sim\gauss(0,\eye)$ of the same shape as the action chunk, the expert integrates the velocity field $v_\theta(\mathbf{A}^{\tau},\tau,\bo_t)$ over $\tau\in[0,1]$ to obtain $\mathbf{A}_t=(\ba_t,\dots,\ba_{t+H-1})$. Training regresses $v_\theta$ onto the velocity of the straight path between a noise sample and a demonstration chunk \cite{lipman2023flow,black2024pi0}; \ac{SFT} uses the same loss on the target-embodiment demonstrations.
 
\textbf{DSRL as an MDP.} The task is an \ac{MDP} with states $\bs_t\in\states$, actions $\ba_t\in\actions$, reward $\reward_t=R(\bs_t,\ba_t)$, transition kernel $P(\bs_{t+1}\given\bs_t,\ba_t)$ and discount $\discountFactor$, and the objective is $\expectation[\sum_{t}\discountFactor^{t}\reward_t]$ over a horizon $T$. \ac{DSRL} lifts this to the noise space: the steering policy's action is $\mathbf{w}$, the environment transition is the composition of the frozen decoder $f_\theta$ with the executed chunk, and the reward accumulated over the chunk is credited to $\mathbf{w}$. The noise policy is trained with SAC \cite{haarnoja2018sac}; App. \ref{app:dsrl} explains the large-chunk variant and hyperparameters we use.

\section{Recipe implementation on $\pi_{0.5}$}
\label{app:recipes}

\textbf{Parameter groups per recipe.} \Cref{tab:recipes} lists the parameter groups each recipe updates. In \recipeExpert, the trainable set is the action expert's attention and MLP projections, its adaptive-RMSNorm modulation layers, the time-embedding MLPs and the action projections; the vision encoder, multimodal projector, and language model are frozen. \recipeFrozenLM\ additionally trains the SigLIP encoder and the multimodal projector. \recipeFull\ trains every parameter used in action generation. \recipeLoRA\ attaches rank-32 adapters ($\alpha=r$) to all attention and MLP projections of the language model and the expert and to the time MLPs, and trains the vision encoder, multimodal projector and action projections in full; the expert's adaptive RMSNorm layers are frozen without adapters, so under LoRA the timestep pathway can adapt only upstream of the modulation. \recipeLoRALM\ applies the same rank-32 adapters to the language model only, and trains the action expert (including its adaptive-RMSNorm layers), the time MLPs, the vision encoder, the multimodal projector and the action projections in full; it differs from \recipeFrozenLM\ only in adapting the language model.
 
\begin{table}[t]
\centering
\small
\caption{SFT recipes on $\pi_{0.5}$. Trainable counts are parameters that receive a gradient, as a fraction of the 3.24B parameters updated by \recipeFull.\protect\footnotemark\ LoRA counts include adapters and the fully trained groups. Everything not listed as updated is frozen.}
\label{tab:recipes}
\begin{tabular}{@{}l p{4.2cm} p{2.6cm} r p{2.0cm}@{}}
\toprule
Recipe & Updated & Frozen & Trainable & Peak LR \\
\midrule
\recipeExpert & Expert attention, MLP and adaRMS layers; time MLPs; action projections & LM, embeddings, SigLIP, multimodal projector & 430M (13.3\%) & $2.5\times10^{-5}$ \\
\recipeFrozenLM & As above, plus SigLIP and multimodal projector & LM, embeddings & 845M (26.1\%) & $2.5\times10^{-5}$ \\
\recipeFull & All parameters & --- & 3.24B (100\%) & $2.5\times10^{-5}$ \\
\recipeLoRA & LoRA $r{=}32$ on LM and expert attention/MLP projections and time MLPs; SigLIP, multimodal projector, action projections in full & Embeddings, expert adaRMS layers, base weights under LoRA adapters & 468M (14.4\%) & $2.5\times10^{-4}$ LoRA, $2.5\times10^{-5}$ full \\
\recipeLoRALM & LoRA $r{=}32$ on LM attention/MLP projections; expert (incl.\ adaRMS), time MLPs, SigLIP, multimodal projector, action projections in full & Embeddings, base weights under LoRA adapters & 887M (27.4\%) & $2.5\times10^{-4}$ LoRA, $2.5\times10^{-5}$ full \\
\bottomrule
\end{tabular}
\end{table}
\footnotetext{The stored checkpoint holds 4.14B parameters, of which $\sim$0.9B receive no gradient under any recipe because they are unused in action generation once the expert reads the key--value cache: the expert's language-modelling head (263M), the backbone's untied language-modelling head (527M), and the backbone's final-layer MLP tail (109M).}

Additional training details are listed in \Cref{tab:sft-hparams}. We use one SFT seed per recipe and three DSRL seeds per selected checkpoint.

\section{DSRL details}
\label{app:dsrl}

\textbf{Large-chunk steering.} Rather than learning a policy over the full chunk noise $\mathbf{A}^0\in\mathbb{R}^{32\times32}$, the actor outputs a single vector $\mathbf{w}\in[-1,1]^{32}$, which is tiled across the 32 chunk positions to form $\mathbf{A}^0$. Repeating one vector keeps the RL action space at 32 dimensions, but it also means DSRL never queries the model under the i.i.d.\ per-position noise it was trained with; App. \ref{app:eval} describes how we account for this when scoring checkpoints. Each macro step decodes one chunk and executes its first 16 actions before the actor is queried again.
 
\textbf{Actor and critics.} The actor and the two critics share a small convolutional encoder over one third-person camera at $64\times64$ and the proprioceptive state. The frozen $\pi_{0.5}$ receives its own camera inputs through its standard preprocessing.
 
\textbf{Learning.} The learner is SAC \cite{haarnoja2018sac} with a reward of $-1$ per macro step until success and a budget of 500k environment steps per run. Each (recipe, checkpoint, task) configuration is run with three RL seeds; runs saturate by about 225k steps. In-loop evaluation uses the deterministic actor every 25k steps. \Cref{tab:dsrl-hparams} lists all hyperparameters, which are identical across recipes, checkpoints and tasks.

\section{Evaluation protocol}
\label{app:eval}
 
\textbf{Two noise distributions before RL.} At inference, the SFT policy's output is a deterministic function of the observation and the initial noise $\mathbf{A}^0$, so its success rate depends on how $\mathbf{A}^0$ is sampled. The standard evaluation draws i.i.d.\ Gaussian noise for every entry and measures the policy as it would ordinarily be deployed. Because DSRL only ever presents the model with one vector repeated across the chunk, we also evaluate under this \emph{repeated} noise, which measures the marginal success rate over the space the actor searches. We report both; the repeated-noise number is the default ``before RL'' figure in the main text.
 
\textbf{Checkpoint selection practice.} We steer four checkpoints per recipe and task instead of pre-selecting one through a heuristic.
 
\textbf{Episode counts.} Before RL, every placement is scored with 25 episodes under each noise distribution, except the two extrapolation placements of \textsc{ScrewLightbulb}, which use 20. After RL, the training placements of both tasks and the extrapolation placements of \textsc{CoffeePressButton} are scored in the loop: five episodes per placement every 25k environment steps, averaged over the evaluations between 400k and 500k steps. The extrapolation placements of \textsc{ScrewLightbulb} were added after those runs completed and are scored by re-evaluating each run's final actor for 10 episodes per placement. All post-RL numbers use the deterministic actor and are averaged over the three RL seeds.

\section{Tasks and demonstration data}
\label{app:tasks}
 
Both tasks replace RoboCasa's parallel-jaw gripper with the LEAP hand. They differ in arm, action space, and control rate. Demonstrations were collected by teleoperation in simulation with a custom setup inspired by ARCap \cite{arcap2025}, using the Meta Quest 3 headset for the arm pose and the Rokoko Smartglove for the finger joint positions.

\section{Full results}
\label{app:full-results}
 
\begin{figure}[ht]
\centering
\includegraphics[width=\linewidth]{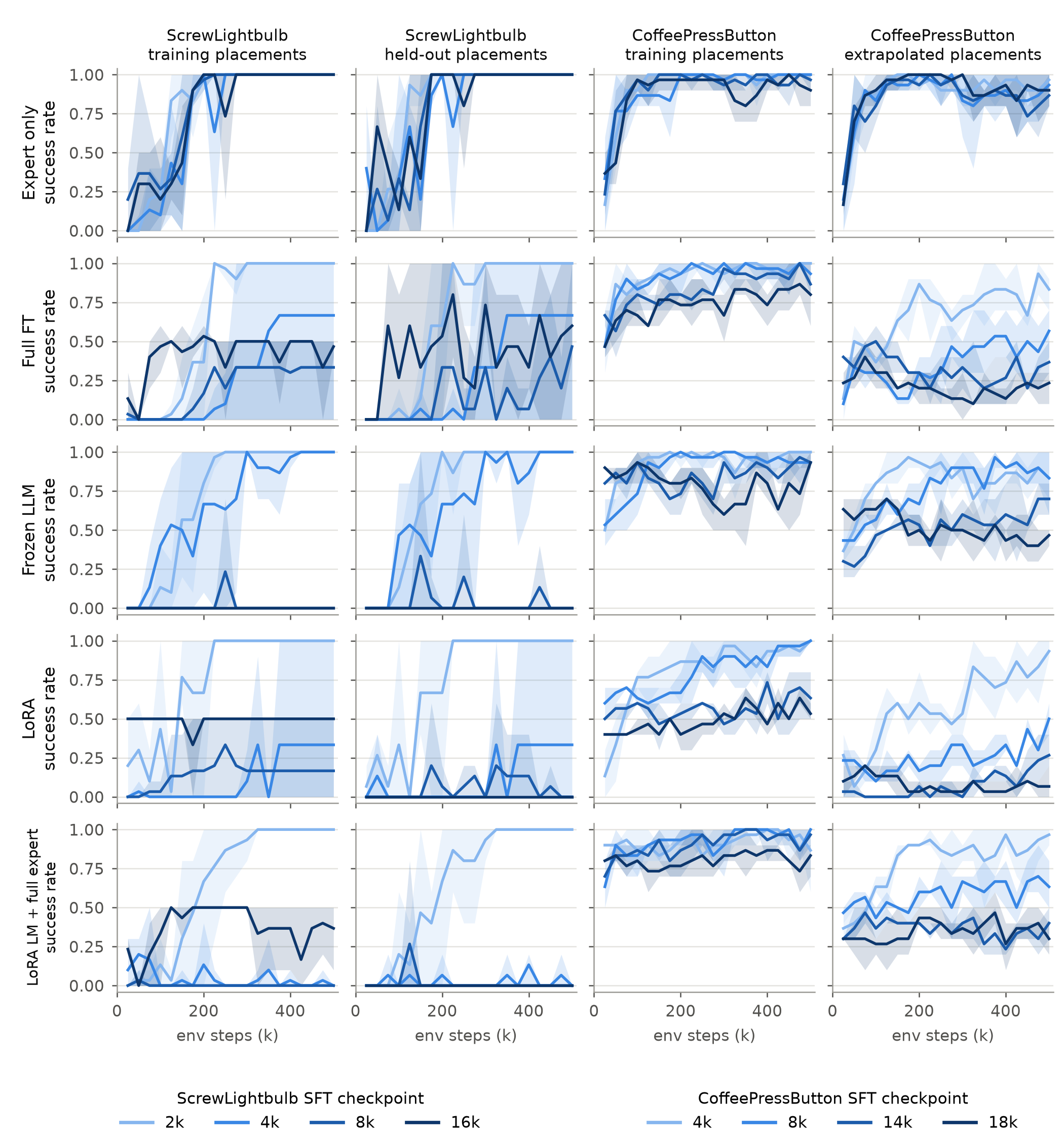}
\caption{\textbf{DSRL learning curves.} We report the in-loop success rate on training placements versus environment steps, averaged across three RL seeds. Training consistently saturates by approximately 225k steps across the 500k-step budget.}
\label{fig:curves}
\end{figure}

\begin{figure}[ht]
\centering
\includegraphics[width=\linewidth]{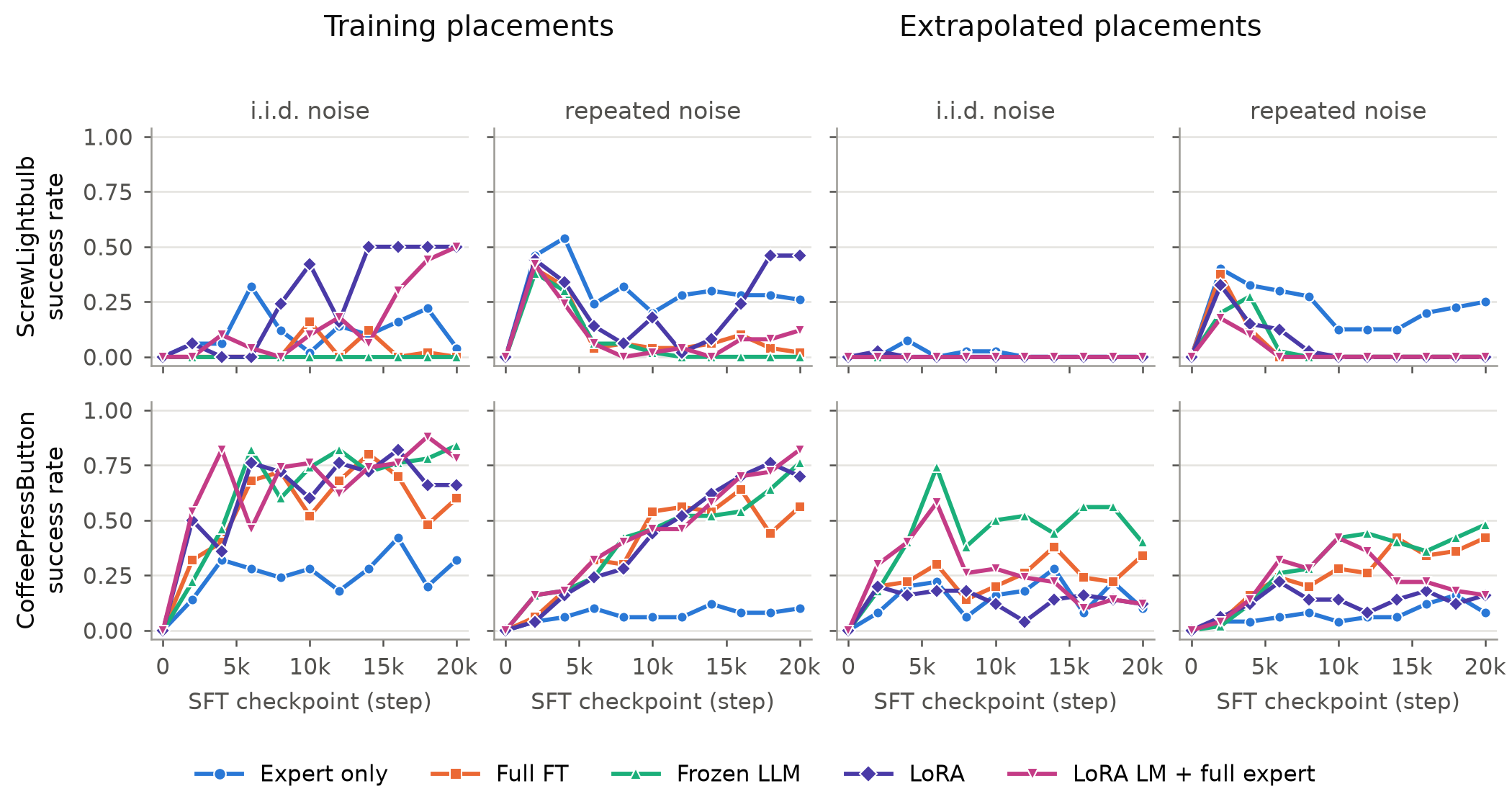}
\caption{\textbf{SFT-time success under independent and repeated noise distributions.} Success rates before RL are shown under i.i.d. noise (standard deployment) and repeated noise (the space searched by DSRL) across SFT steps and recipes. The two measures fail to reliably track post-DSRL success.}
\label{fig:before_rl}
\end{figure}

\Cref{tab:main} lists every configuration in the study, including SFT-time success under repeated noise. Repeated-noise success is not much better than i.i.d. noise for predicting steerability: \recipeExpert\ remains at 6--12\% at every checkpoint and steers to 95--100\%, while \recipeLoRA\ at 18k has the highest repeated-noise success of any configuration (76\%) and the lowest post-DSRL success. \Cref{fig:curves} shows the DSRL learning curves and \Cref{fig:before_rl} shows the pre-DSRL checkpoint performance under the two noise distributions.

\begin{table}[t]
\centering
\small
\caption{Tasks and demonstration data. Both tasks use a 16-DoF LEAP hand.}
\label{tab:tasks}
\begin{tabular}{@{}lllrrr@{}}
\toprule
Task & Arm & Action representation & Rate & Demos & Frames \\
\midrule
\textsc{CoffeePressButton} & Franka Panda & EE pose deltas (OSC), 6 + 16 & \SI{20}{Hz} & 56 & 7{,}928 \\
\textsc{ScrewLightbulb} & xArm6 & absolute joint targets, 6 + 16 & \SI{30}{Hz} & 33 & 31{,}842 \\
\bottomrule
\end{tabular}
\end{table}

\begin{table}[t]
\centering
\small
\caption{SFT hyperparameters.}
\label{tab:sft-hparams}
\begin{tabular}{@{}ll@{}}
\toprule
Hyperparameter & Value \\
\midrule
Train steps & 20k \\
Batch size & 32 \\
Optimizer & AdamW, $\beta=(0.9,0.95)$, weight decay 0.01 \\
Gradient clipping & 1.0 \\
LR schedule & 1k linear warmup, cosine decay to $0.1\times$ peak \\
Peak LR, fully trained parameters & $2.5\times10^{-5}$ \\
Peak LR, LoRA parameters (\recipeLoRA, \recipeLoRALM) & $2.5\times10^{-4}$ \\
LoRA rank / $\alpha$ (\recipeLoRA, \recipeLoRALM) & 32 / 32 \\
Action chunk & $H=32$ predicted, 16 executed \\
Image augmentation & brightness 0.3, contrast 0.4, saturation 0.5, hue 0.08. \\
Proprioceptive-state dropout & $p=0.2$ \\
Checkpoint interval & 2k steps \\
SFT seeds per recipe & 1 \\
DSRL seeds per selected checkpoint & 3 \\
\bottomrule
\end{tabular}
\end{table}

\begin{table}[ht]
\centering
\small
\caption{DSRL hyperparameters.}
\label{tab:dsrl-hparams}
\begin{tabular}{@{}ll@{}}
\toprule
Hyperparameter & Value \\
\midrule
Noise actor & single-step, $\mathbf{w}\in[-1,1]^{32}$ ($\tanh$), Repeated over $H=32$ \\
Actions executed per macro step & 16 of 32 (\SI{0.8}{s} at \SI{20}{Hz}; \SI{0.53}{s} at \SI{30}{Hz}) \\
Actor/critic inputs & 1 third-person image ($64\times64$) + Proprioceptive state \\
Encoder & small CNN; 64-d image latent, 64-d state latent \\
Learner & SAC, 10 $Q$-heads reduced by mean \\
Update-to-data ratio & 20 \\
Batch size & 256 \\
Replay buffer & 100k macro transitions \\
Replay seeding & 1{,}000 macro transitions with $\mathbf{A}^0\sim\gauss(0,\eye)$ \\
Learning rate & $3\times10^{-4}$ \\
Target entropy & $-16$ (half the noise dimension) \\
Reward & $-1$ per macro step until success (terminal) \\
Discount & 0.999 per primitive step ($\approx0.984$ per macro step) \\
Budget & 500k primitive environment steps \\
RL seeds & 3 (actor/critic init, exploration noise, replay order) \\
\bottomrule
\end{tabular}
\end{table}

\begin{table}[t]
\centering
\small
\setlength{\tabcolsep}{5pt}
\caption{\textbf{Success rate (\%) before and after steering, for every recipe and checkpoint.} Before RL: training placements under i.i.d.\ and repeated noise, and extrapolation placements under i.i.d.\ noise. After DSRL: training and extrapolation placements, mean over three RL seeds. Bold marks each recipe's best post-DSRL extrapolation checkpoint.}
\label{tab:main}
\begin{tabular}{@{}ll rrr rr@{}}
\toprule
 & & \multicolumn{3}{c}{Before RL} & \multicolumn{2}{c}{After DSRL} \\
\cmidrule(lr){3-5}\cmidrule(l){6-7}
Recipe & SFT step & train (i.i.d.) & train (rep.) & extrap.\ (i.i.d.) & train & extrap. \\
\midrule
\multicolumn{7}{@{}l}{\textsc{ScrewLightbulb}}\\ \midrule
\recipeExpert & 2k & 6 & 46 & 0 & 100 & \textbf{98} \\
 & 4k & 6 & 54 & 8 & 100 & 73 \\
 & 8k & 12 & 32 & 2 & 100 & 88 \\
 & 16k & 16 & 28 & 0 & 100 & 45 \\
\recipeFrozenLM & 2k & 0 & 38 & 0 & 100 & \textbf{87} \\
 & 4k & 0 & 30 & 0 & 99 & 70 \\
 & 8k & 0 & 6 & 0 & 0 & 0 \\
 & 16k & 0 & 0 & 0 & 0 & 0 \\
\recipeFull & 2k & 0 & 40 & 0 & 100 & \textbf{73} \\
 & 4k & 0 & 32 & 0 & 67 & 15 \\
 & 8k & 0 & 6 & 0 & 33 & 0 \\
 & 16k & 0 & 10 & 0 & 46 & 0 \\
\recipeLoRA & 2k & 6 & 44 & 2 & 100 & \textbf{63} \\
 & 4k & 0 & 34 & 0 & 33 & 5 \\
 & 8k & 24 & 6 & 0 & 17 & 0 \\
 & 16k & 50 & 24 & 0 & 50 & 0 \\
 \recipeLoRALM & 2k & 0 & 42 & 0 & 100 & \textbf{85} \\
 & 4k & 10 & 24 & 0 & 2 & 3 \\
 & 8k & 0 & 0 & 0 & 0 & 0 \\
 & 16k & 30 & 8 & 0 & 38 & 0 \\
\midrule
\multicolumn{7}{@{}l}{\textsc{CoffeePressButton}}\\ \midrule
\recipeExpert & 4k & 32 & 6 & 20 & 100 & \textbf{94} \\
 & 8k & 24 & 6 & 6 & 99 & 87 \\
 & 14k & 28 & 12 & 28 & 97 & 83 \\
 & 18k & 20 & 8 & 22 & 95 & 90 \\
\recipeFrozenLM & 4k & 46 & 18 & 40 & 98 & 85 \\
 & 8k & 60 & 42 & 38 & 94 & \textbf{89} \\
 & 14k & 72 & 52 & 44 & 91 & 62 \\
 & 18k & 78 & 64 & 56 & 78 & 43 \\
\recipeFull & 4k & 40 & 18 & 22 & 97 & \textbf{81} \\
 & 8k & 72 & 30 & 14 & 96 & 49 \\
 & 14k & 80 & 54 & 38 & 93 & 31 \\
 & 18k & 48 & 44 & 22 & 81 & 20 \\
\recipeLoRA & 4k & 36 & 16 & 16 & 95 & \textbf{83} \\
 & 8k & 72 & 28 & 18 & 95 & 35 \\
 & 14k & 72 & 62 & 14 & 65 & 17 \\
 & 18k & 66 & 76 & 14 & 55 & 7 \\
 \recipeLoRALM & 4k & 82 & 18 & 40 & 95 & \textbf{91} \\
 & 8k & 74 & 40 & 26 & 97 & 63 \\
 & 14k & 74 & 58 & 22 & 94 & 33 \\
 & 18k & 88 & 72 & 14 & 82 & 34 \\
\bottomrule
\end{tabular}
\end{table}

\begin{figure}[ht]
\centering
\includegraphics[width=\linewidth]{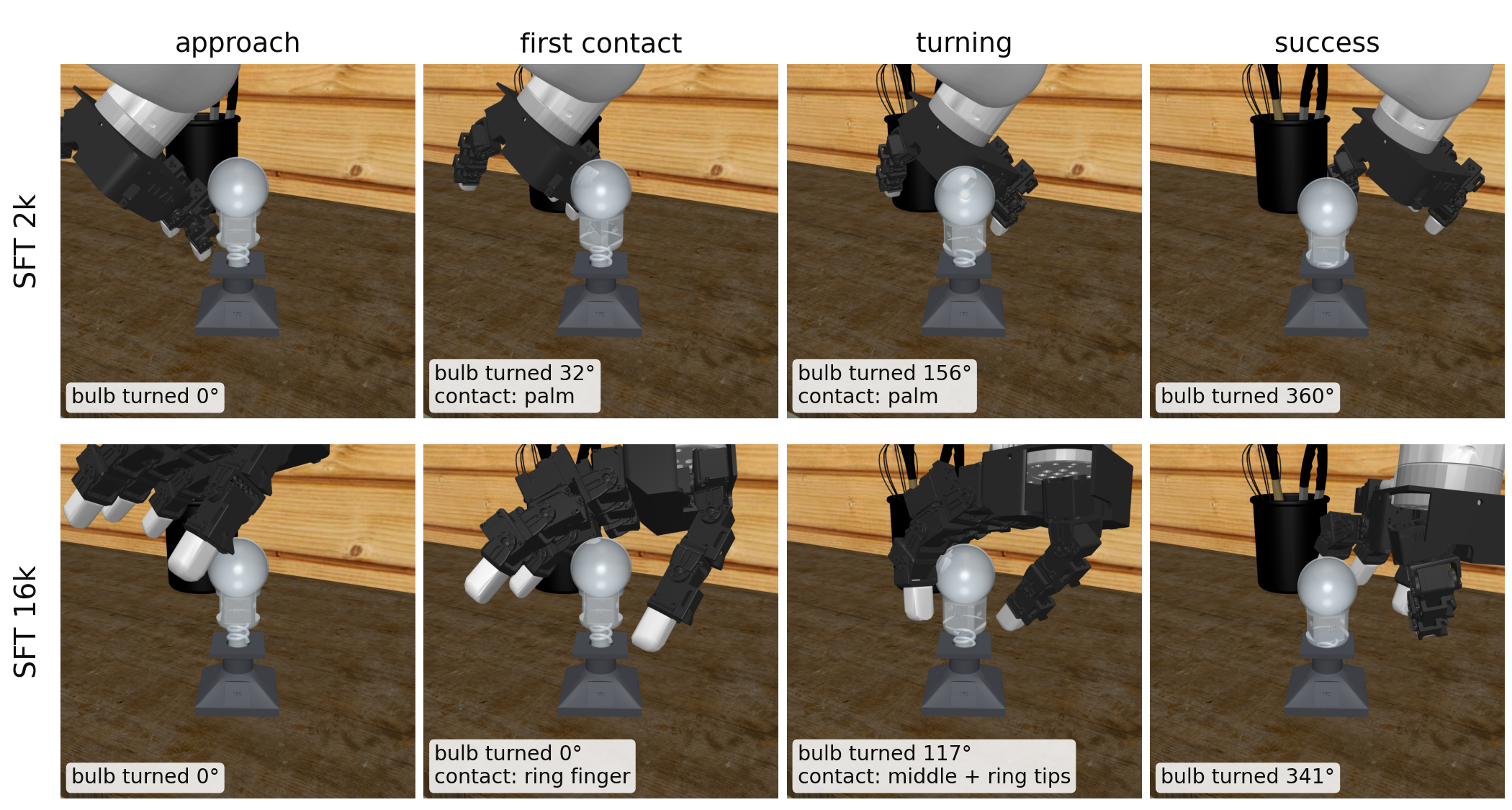}
\caption{\textbf{Qualitative rollouts demonstrating the evolution of skill acquisition along SFT checkpoints.} Steered \recipeExpert\ execution on the \textsc{ScrewLightbulb} task reveals differing strategies. Top: Early checkpoints (2k steps) exploit the simulator's success condition by using the palm without a stable grasp. Bottom: Later checkpoints (16k steps) successfully reproduce the demonstrated dexterous behavior, exhibiting a stable grasp and twist.}
\label{fig:filmstrip}
\end{figure}

\end{appendices}

\end{document}